\documentclass{article}
\usepackage{caption}
\usepackage[utf8]{inputenc}

\usepackage{multirow}
\usepackage{diagbox}

\usepackage[preprint]{neurips_2025}
\usepackage[table,xcdraw]{xcolor}
\usepackage{booktabs} 
\usepackage{multirow} 

\usepackage{amsmath}
\usepackage{amssymb}

\usepackage{wrapfig}

\usepackage{graphicx}
\usepackage[utf8]{inputenc} 
\usepackage[T1]{fontenc}    
\usepackage{hyperref}       
\usepackage{url}            
\usepackage{booktabs}       
\usepackage{amsfonts}       
\usepackage{nicefrac}       
\usepackage{microtype}      
\usepackage{xcolor}         
\usepackage{natbib}
\usepackage{algorithm}
\usepackage{algpseudocode}
\usepackage{listings}
\usepackage{xcolor}

\usepackage{algorithmicx}
\usepackage{amsmath,amssymb}
\usepackage{makecell}
\usepackage{tabularray}
\usepackage{float}

\usepackage{authblk}
\usepackage{hyperref}
\newcommand{\Input}{\item[\textbf{Input:}]}
\newcommand{\Output}{\item[\textbf{Output:}]}

\title{DHEvo: Data-Algorithm Based Heuristic Evolution for Generalizable MILP Solving}

\author[1]{Zhihao Zhang}
\author[1,3]{Siyuan Li}
\author[1]{Chenxi Li}
\author[1]{Feifan Liu}
\author[2]{Mengjing Chen}
\author[2]{Kai Li}
\author[2]{Tao Zhong}
\author[3]{Bo An}
\author[1]{Peng Liu}

\affil[1]{Harbin Institute of Technology}
\affil[2]{Huawei Noah’s Ark Lab}
\affil[3]{Nanyang Technological University}

\begin{document}

\maketitle

\begin{abstract}

Primal heuristics play a critical role in improving the efficiency of mixed integer programming (MILP) solvers. As large language models (LLMs) have demonstrated superior code generation abilities, recent MILP works are devoted to leveraging the evolutionary computation approaches with LLMs to generate effective primal heuristics. Although the generated heuristics have achieved better solving performance than the hand-crafted ones with little adaptability, the advantage of current LLM-based methods is limited to few MILP instances in one problem class, as they fail to capture the instance characteristics in the problem class (the MILP instances generated from the same mathematical model are defined as a problem class). Since MILP instances often differ significantly in structure and feature distribution, the neglect of their characteristics in the evolution process results in poor generalization within the same problem class.
To overcome this challenge, we propose a data-algorithm co-evolution framework (DHEvo) that iteratively selects representative instances and evolves corresponding heuristics. 
With the initial instance distribution, we develop an LLM-based multi-agent system to generate data-code pairs simultaneously.
These data-code pairs are iteratively refined based on their fitness scores, leading to the identification of the most effective heuristic over the entire problem class. 
Extensive experiments across diverse MILP benchmarks demonstrate that our approach significantly outperforms both human-designed heuristics and existing LLM-based methods.

\end{abstract}

\section{Introduction}

Mixed-Integer Linear Programming (MILP) constitutes a fundamental modeling and solution framework in operations research. It has been widely applied to a broad range of real-world problems, including supply chain optimization~\citep{liu2008tsp,jeong2019biodiesel,jokinen2015milp}, hardware design~\citep{ma2019accelerating,hafer1991constraint}, production scheduling~\citep{chen2010integrated,caumond2009milp,superchi2024optimization}, and energy management~\citep{chang2004practical,kassab2024optimal,zare2024efficient}. In practical applications, a complex MILP problem is often defined by numerous parameters, such as cost coefficients, constraints, and bounds. These can all be mathematically represented as:
\[
z^{\dagger} := \min_{x \in P^{\dagger}} c^{\top}x, \quad P^{\dagger} = \left\{ x \in \mathbb{R}^{n} \mid Ax < b, \underline{\pi} \leq x \leq \overline{\pi}, x_{j} \in \mathbb{Z} \ \forall j \in \mathcal{I} \right\},
\]
where \( M^{\dagger} := (c, P^{\dagger}) \), \( A \in \mathbb{R}^{m \times n} \), \( b \in \mathbb{R}^{m} \), \( c, x \in \mathbb{R}^{n} \), \( \underline{\pi}, \overline{\pi} \in \mathbb{R}_{\infty}^{n} \), and \( \mathcal{I} \subseteq \{1, \ldots, n\} \) indexes the integer-constrained variables.

In real-world settings, each instance corresponds to a unique and specific problem configuration, even if it originates from the same MILP model. Moreover, the structural and statistical characteristics of these instances can vary significantly, leading to large intra-class diversity.  Therefore, well-designed primal heuristics ~\citep{wong1984worst,balas2004pivot,berthold2006primal,wallace2010zi,witzig2021conflict}, which aim to find feasible solutions quickly, are important not only for improving solver efficiency but also for achieving robust generalization across instances within the same problem class.


Current heuristic design approaches for MILP can be broadly categorized into two types: (i) manually designed and (ii) automatically generated in an adaptive manner. Manually designed heuristics, such as those implemented in commercial solvers like SCIP  \citep{scip} and Gurobi  \citep{gurobi}, rely heavily on expert knowledge and often struggle to adapt to novel problem instances.
The alternative paradigm is the dynamic generation of heuristic algorithms. It mainly combines large language models (LLMs) with evolutionary computation (EC) to automate the design of heuristic algorithms  \citep{yang2023large,meyerson2024language,chen2023evoprompting,romera2024mathematical,liu2024systematic,liu2024evolution,zhou2024llm4solver}. These approaches aim to leverage the reasoning capabilities of LLMs to generate more effective and adaptive heuristics. Nevertheless, current methods mostly perform the evolutionary computation process on only a few instances. Consequently, the generated algorithms fail to capture the shared instance characteristics in the problem class, leading to insufficient representational ability and poor generalization across instances within the same problem class, which can be observed in our ablation study. In summary, both manually crafted and automatically generated heuristics struggle to balance efficiency with generalization. This stems from their limited ability to model the instance distribution and capture shared structural patterns across different instances.

To address this, we propose a data-algorithm co-evolution framework (DHEvo) that generates generalizable algorithms by iteratively evolving both the MILP data and the algorithms. We start by randomly sampling instances from a domain-specific dataset and using an LLM-based multi-agent evolution system (MA-Evolution System) to create initial data-code pairs. Inspired by few-shot  \citep{fewshot1,fewshot2,fewshot3,fewshot4}, we assume that data-code pairs with the highest fitness (relative primal gap) are more likely to generalize well across instances within a problem class. Therefore, the pairs with the highest fitness scores are selected to form the initial population for further evolution. This process is iterated over multiple generations, gradually narrowing down to the most representative algorithm and instance.
\begin{figure}
  \centering
  \small
\includegraphics[width=1\textwidth]{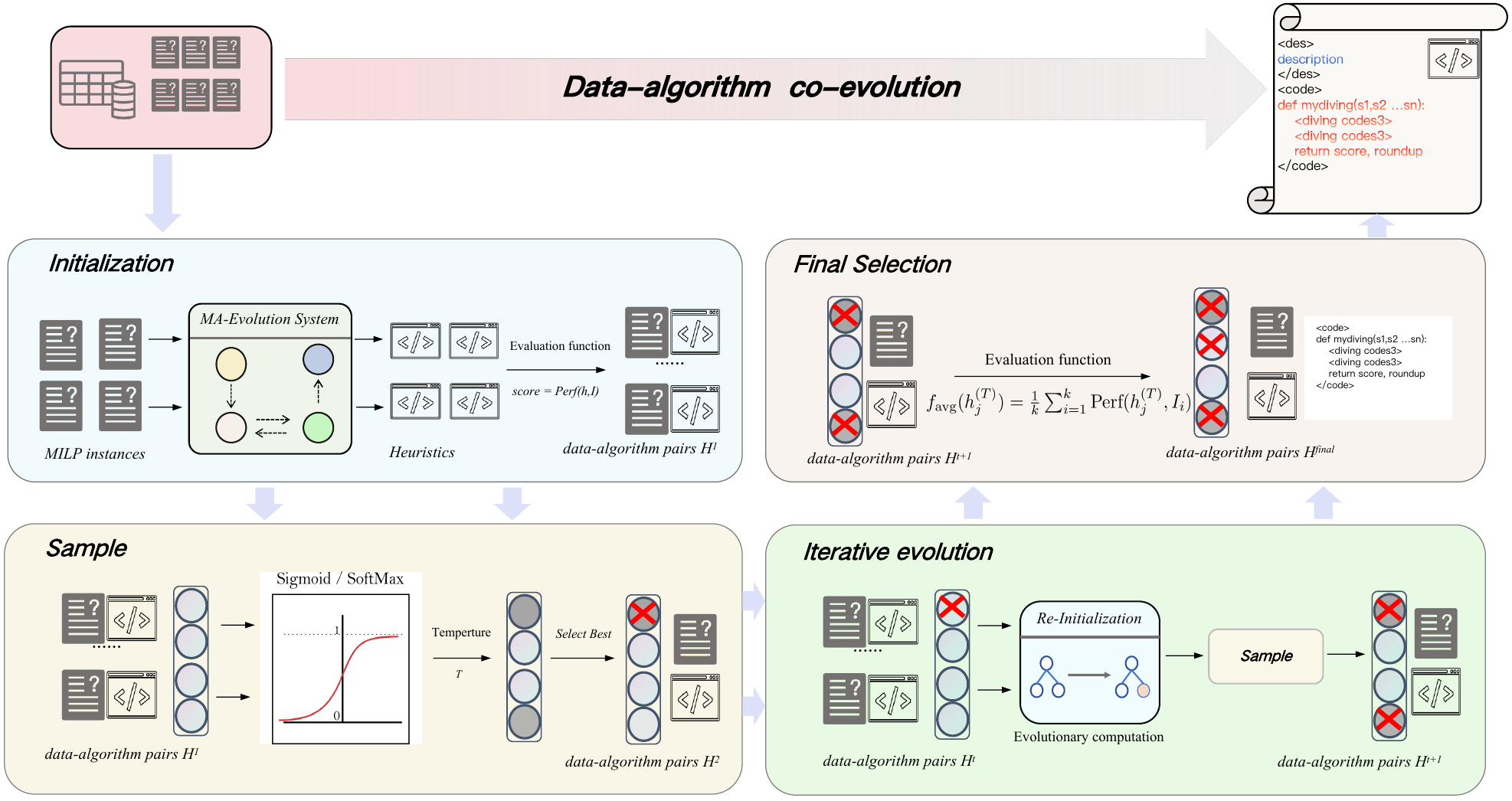}
    \caption{Illustration of data-algorithm co-evolution framework (DHEvo).}
  \label{main-method}

\end{figure}
In summary, our contributions are as follows:
\begin{itemize}
\item We propose a unified evolutionary computation framework based on data-algorithm co-evolution. It enables better approximation of the instance distribution and enhances the representational capacity of the learned heuristics, leading to improved generalization.
    
\item We further introduce a co-evolutionary solution tailored to MILP tasks by incorporating a multi-agent evolution system (MA-Evolution System). It refines the evolutionary process, minimizing generation errors and optimizing the quality of the resulting heuristics.
    \item Extensive experiments show that our method significantly improves the generalization of diving heuristics and delivers substantial performance gains across multiple MILP datasets.
\end{itemize}

\section{Background and related works}
\label{Background}

\subsection{Branch\&Bound and diving heuristic}

A common approach to solving MILP problems is Branch-and-Bound (B\&B)  \citep{land2009automatic}, which recursively builds a search tree by selecting variables and partitioning the problem into subproblems. These subproblems are created by adding constraints based on the fractional value of the variable in the LP relaxation solution. B\&B also prunes branches using objective bounds to improve efficiency. However, B\&B can be computationally expensive for large-scale problems, so primal heuristics like the diving heuristic are often used to accelerate the search. Diving performs a depth-first search by iteratively rounding variables and solving the linear program until a feasible solution is found or infeasibility is proven. While existing diving heuristics are effective, they often require manual tuning and expert knowledge to design. In contrast, our approach uses evolutionary computation to automatically generate problem-specific heuristics, offering more flexibility, adaptability, and reduced reliance on expert knowledge. Our experiments show that this approach significantly improves solver performance across multiple datasets.

\subsection{Performance measurement}
To evaluate the performance of MILP solvers, we use several key performance metrics: Primal-Dual Gap, Primal-Dual Integral, and Primal Gap.

\textbf{Primal-Dual Gap} It is a widely used measure that quantifies the difference between the primal objective value and the dual objective value at any given time during the optimization process. It gives an indication of how close the current solution $\tilde{z}$ is to an optimal solution $\tilde{z}^*$. Mathematically, the Primal-Dual Gap is defined as:
\[
\gamma_{pd}(\tilde{z}, \tilde{z}^*) =
\begin{cases}
\frac{|\tilde{z} - \tilde{z}^*|}{\max(|\tilde{z}|, |\tilde{z}^*|)} & \text{if } 0 < \tilde{z}, \tilde{z}^* < \infty, \\
1 & \text{otherwise}.
\end{cases}
\]

\textbf{Primal-Dual Integral} While the primal-dual gap captures a snapshot at a particular time, the primal-dual integral evaluates the solver's progress over the entire solving process by aggregating the primal-dual gap over time. It is given by:
\[
\gamma_{pdi}(t) = \int_0^t \gamma_{pd}(\tilde{z}(\tau), \tilde{z}^*(\tau)) \, d\tau,
\]
where \(\gamma_{pd}(\tilde{z}(\tau), \tilde{z}^*(\tau))\) represents the Primal-Dual Gap at time \(\tau\).


\textbf{Primal Gap} It is used to evaluate the effectiveness of diving heuristics, which primarily aim to improve the primal performance by guiding the search toward better feasible solutions. The relative primal gap is defined as the absolute difference between the current objective value \(\tilde{z}\) and the optimal solution \(z^{\dagger}\), normalized by the objective value of the optimal solution. The formula for the primal gap is given by:
\[
\gamma_p(\tilde{z}) = \frac{|\tilde{z} - z^{\dagger}|}{|z^{\dagger}|},
\]

where \(z^{\dagger}\) is the objective value of the optimal solution obtained after presolving.
In the case where \(|z^{\dagger}| = 0\), we use the following modified primal gap:
\[
\gamma'_p(\tilde{z}) = |\tilde{z} - z^{\dagger}|.
\]

\subsection{LLM for evolutionary computation}
Evolutionary computation is a widely used method for solving optimization problems inspired by natural evolution  \citep{back1997handbook}. Given an optimization problem, evolutionary algorithms  \citep{zhou2019evolutionary,eiben2015evolutionary} treat each candidate solution as an individual. They iteratively apply operations such as parent selection, crossover, mutation, fitness evaluation, and survivor selection to evolve the population toward better solutions. 

In recent years, the capabilities of large language models have advanced significantly  \citep{naveed2023comprehensive}. Researchers have started exploring the integration of LLMs into evolutionary computation frameworks to automatically generate heuristic algorithms  \citep{liu2024systematic,zhang2024understanding,wu2024evolutionary}. For example, Funsearch  \citep{romera2024mathematical} has developed an evolutionary computation framework based on LLM to improve the quality of generated functions iteratively. EoH  \citep{liu2024evolution} further integrates thoughts with code to generate more effective algorithms and has achieved promising results, such as the online bin packing problem. Current methods typically train on a limited set of specific instances. This prevents large language models from learning the common characteristics of the problem class. As a result, the generated algorithms excel in similar instances but lack generalization with others within the same class.

\section{Method}
In this section, we introduce the problem setting and our key insights in \autoref{sec1}, and then present the proposed data-algorithm co-evolution framework in \autoref{sec2}. 
Finally, we detail the implementation process, including the specific evolutionary operation and the design of the MA-Evolution System in \autoref{sec3}.

\subsection{Problem formulation and insight}
\label{sec1}

\begin{figure}
  \centering
  \small
\includegraphics[width=0.75\textwidth]{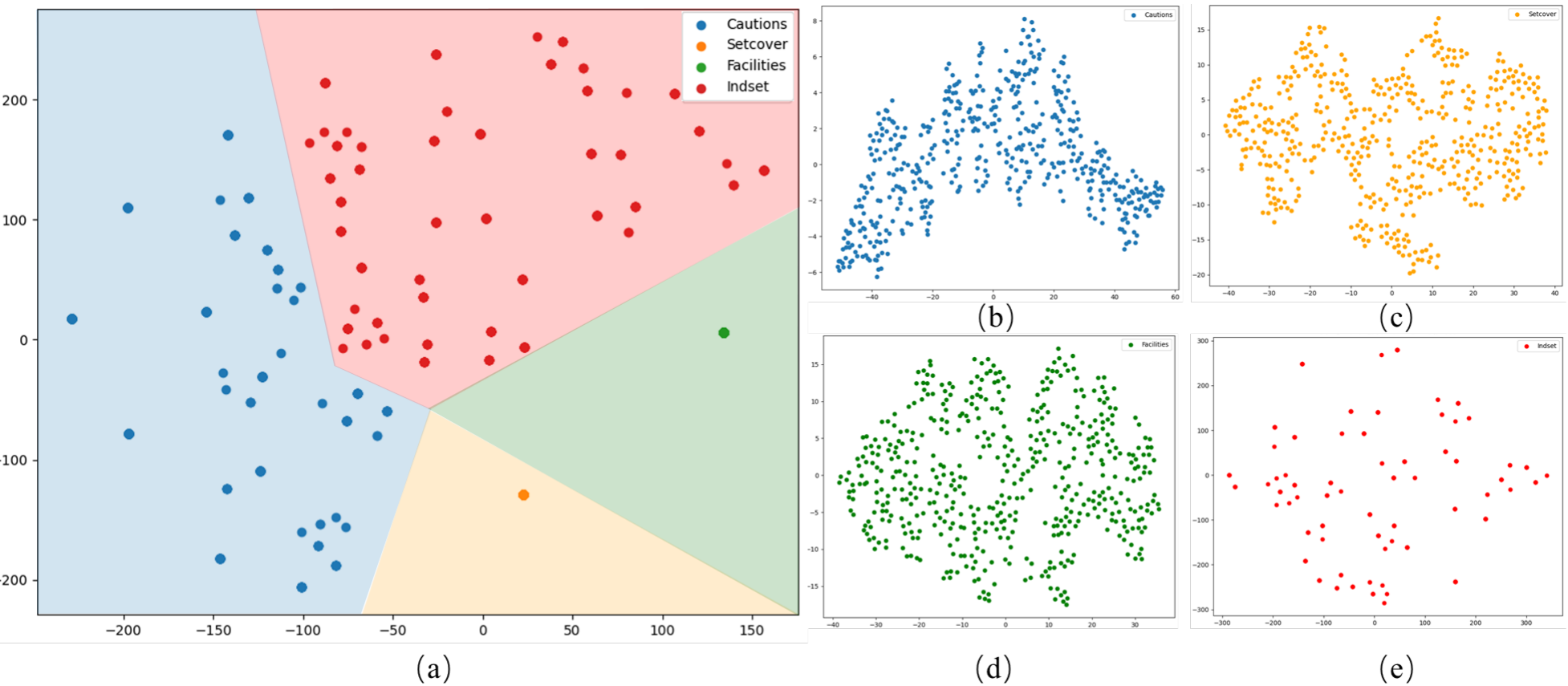}
  \caption{The visualization of instance features via t-SNE is presented as follows: Panel a represents the overall distribution of instances across four datasets, while panels b, c, d, and e show the distributions for Cauctions, Setcover, Facilities, and Indset, respectively. The Setcover (orange) and Facilities (green) datasets are represented as single points due to their fixed numbers and proportions of variables and constraints. }
  \vspace{-3mm}
  \label{fig:instance}
\end{figure}

In real-world MILP tasks, instances within the same problem class can differ significantly in their distributions, constraints, and structures. However, they often share common features, such as similar types of constraints, variable ranges, or recurring patterns in the objective function. As shown in \autoref{fig:instance}, we visualize 17 typical features from the four combinatorial optimization datasets, including variable-constraint ratios (detailed results can be found in the appendix). How to extract these shared features within the same problem class to improve algorithm performance has become a key focus in large language model evolutionary computation research.

Current evolutionary computation methods typically evaluate heuristics by averaging their performance over a small number of randomly selected instances. Specifically, given a heuristic algorithm \( h \) and a set of instances \( \mathcal{I} = \{I_1, I_2, \dots, I_k\} \), the fitness of the heuristic is calculated as:
\[
\mathcal{F}(h) = \frac{1}{k} \sum_{i=1}^{k} \text{Perf}(h, I_i),
\]
where \( \text{Perf}(h, I_i) \) represents the performance of the heuristic \( h \) on instance \( I_i \).

This method assumes that all instances are equally representative, which is often not the case in MILP problems, where instances' structure can vary significantly.
To account for this, we introduce the variance of performance \( \sigma^2(h) \) over a broader set of instances \( \mathcal{D} \). A high \( \sigma^2(h) \) indicates that the heuristic performs inconsistently across different instances, which suggests poor generalization.
Inspired by few-shot learning  \citep{fewshot1,fewshot2,fewshot3,fewshot4}, extensive research  \citep{wu2017towards,akbari2021does,jiang2019fantastic} has shown that starting with "simple" or "representative" samples in complex datasets often enhances both learning efficiency and generalization. In this context, to identify the most representative instances, we hypothesize that those with higher fitness scores in the evolutionary computation process will likely exhibit greater structural representativeness. The rationale behind this assumption is as follows:

\textbf{Insight 1: High fitness scores indicate structural representativeness.}
Let \( I \) be an MILP instance with LP relaxation \( I_{LP} \), and let \( \Delta(I) = |z^*_{LP} - z^*| \) denote the integrality gap of \( I \), where \( z^*_{LP} \) and \( z^* \) represent the optimal solutions to the LP relaxation and the integer program, respectively. If the integrality gap is small, \emph{i.e.}, \( \Delta(I) \leq \epsilon \), then heuristics trained on \( I \) will achieve lower variance \( \sigma^2(h) \) to instances with similar LP relaxation tightness \( \Delta(I)\). This suggests that higher fitness scores, which often correspond to tighter LP relaxations, indicate a more representative structure that enables better generalization across similar instances.


\textbf{Insight 2: regular feasible regions enhance heuristic stability.}
Let \( \mathcal{R}(I) \) represent the feasible region of instance \( I \), and \( \nu(I) \) be a measure of the irregularity of \( \mathcal{R}(I) \), such as the number of disconnected components or redundant constraints. If \( \nu(I) \) is small, then heuristics trained on \( I \) will perform more consistently on similar instances with small \( \nu(I) \). This suggests that instances with more regular feasible regions are more stable and, thus, their heuristics are likely to generalize better.


\subsection{Data-Algorithm based heuristic evolution framework }
\label{sec2}
As illustrated in \autoref{main-method}, our framework adopts a structured evolutionary process that tightly couples instance selection with heuristic generation and optimization. This process is designed to progressively improve both the solution quality and the generalization capability of the generated heuristics through data-algorithm co-evolution.

\begin{algorithm}[ht]
\caption{DHEvo framework}
\label{alg:co-evolution}
\begin{algorithmic}[1]

\small
\Require Problem distribution \( \mathcal{D} \), population size \( m \), number of instances \( n \), top-\( k \), total iterations \( T \)
\Ensure Final heuristic algorithm population \( \mathcal{H}^{\text{final}} \)
\State \textbf{Initialization:} Sample initial instance set \( \mathcal{I}_0 \in \mathcal \{I_1, \dots, I_n\} \sim \mathcal{D} \)
\State Generate initial algorithm population \( \mathcal{H}_0 = \{h_1^{(0)}, \dots, h_m^{(0)}\} \) via MA-Evolution System (MA-E)
\For{each \( h_j^{(0)} \in \mathcal{H}_0 \)}
    \State Evaluate \( f(h_j^{(0)}) = \text{Perf}(h_j^{(0)}, I_j) \)
\EndFor

\For{each \( I_i \in \mathcal{I}_0 \)}
    \State Generate \( \mathcal{H}_i^{(1)} = \text{MA-E}(I_i, \mathcal{H}_0) \)
    \State Evaluate fitness for each \( h \in \mathcal{H}_i^{(1)} \)
    \State Select top performer \( h_i^* = \arg\max_{h} \text{Perf}(h, I_i) \)
\EndFor
\State Let \( \mathcal{P}^{*} \leftarrow \text{top-}k \text{ pairs in } \mathcal{P}^{(1)} \text{ ranked by } \operatorname{Perf}(I_i, h_i^*)\)

\For{iteration \( t = 2 \) to \( T \)}
    \State \textbf{Re-Initialization:}
    \For{each \( (I_j, h_j^*) \in \mathcal{P}^* \)}
        \State Generate new candidates via prompt: \( \mathcal{H}_j^{(t)} = \text{MA-E}(I_j, \text{Prompt}(h_j^*)) \)
        \State Evaluate each \( h \in \mathcal{H}_j^{(t)} \) and select \( h_j^{(t)} = \arg\max_{h} \text{Perf}(h, I_j) \)
    \EndFor
    \State Update \( \mathcal{P}^{(t)} = \{(I_j, h_j^{(t)})\} \), smale top-\( k \) pairs \( \mathcal{P}^* \leftarrow Smaple( \text{Perf}(h, I_i)) \)
\EndFor

\State \textbf{Final Selection:}
\State Compute \( f_{\text{avg}}(h_j^{(T)}) = \frac{1}{k} \sum_{i=1}^{k} \text{Perf}(h_j^{(T)}, I_i) \)
\State Select top heuristics \( \mathcal{H}^{\text{final}} = \{h_j^{(T)}\}_{j=1}^k \)
\State \Return \( \mathcal{H}^{\text{final}} \)
\end{algorithmic}
\end{algorithm}

We assign fitness scores to each instance individually rather than averaging them across the dataset, which allows complex or unrepresentative instances to dominate the optimization process and ultimately undermines the generalization ability of the learned heuristics.
As shown in Algorithm \ref{alg:co-evolution}, it begins by selecting a subset of high-quality instance-heuristic pairs \( (\mathcal{I}^*, \mathcal{H}^*) \), where each heuristic in \( \mathcal{H}^* \) demonstrates strong performance on its corresponding instance in \( \mathcal{I}^* \). These pairs are identified by evaluating all candidate heuristics against a sampled MILP dataset and ranking them based on a independent fitness score. These instances with higher scores indicate better structural representativeness for the problem class.

To balance exploration and exploitation during selection, we employ a temperature-controlled strategy. The retention probability \( \mathcal{S} \) for each candidate is modeled by a sigmoid function with temperature parameter \( T \):
\[
\mathcal{S} = \frac{1}{1 + \exp\left(-\frac{1}{T}\right)}.
\]
Higher temperatures encourage diversity by favoring a broader range of candidates, while lower temperatures promote exploitation of already promising heuristics. Once the top-performing pairs \( (\mathcal{I}^*, \mathcal{H}^*) \) are selected, the heuristics in \( \mathcal{H}^* \) are used to extract prompts for generating new candidates via our MA-Evolution System, a multi-agent LLM-based generation and evaluation module. These new heuristics are then re-evaluated on the structurally representative instances from \( \mathcal{I}^* \), ensuring alignment with the overall problem distribution.

This cycle—comprising heuristic generation, evaluation, selection, and re-initialization—is repeated for a fixed number of iterations. Through this iterative process, the framework incrementally converges toward a set of heuristics that exhibit both strong instance-level performance and robust generalization across the problem class.

At the final stage, the top \( k \) instance-heuristic pairs are fixed, and their aggregated performance is assessed. The resulting heuristics constitute the final evolved algorithm portfolio for solving MILP problems.

\begin{figure}
  \centering
  \small
  
\includegraphics[width=1\textwidth]{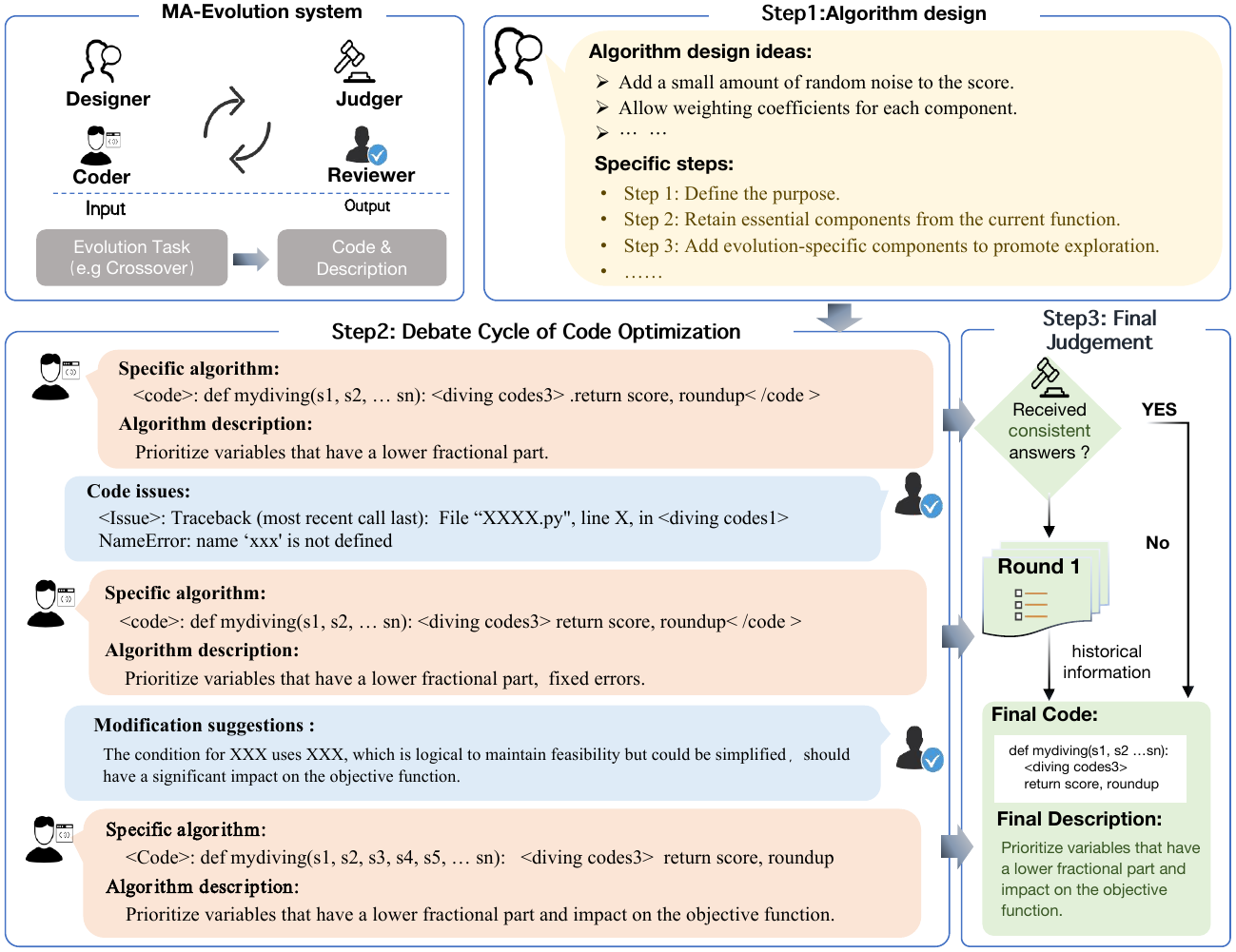}
\vspace{-3mm}
  \caption{Illustration of MA-Evolution System.}
  \label{ma-e}
  \vspace{-3mm}
\end{figure}

\subsection{Framework implementation}
\label{sec3}

\textbf{Evolution operation}
Our evolutionary framework comprises four main operations: initialization, crossover, mutation, and parent selection. As shown in \autoref{ma-e}, we implement initialization, crossover, and mutation through tailored prompts, and leverage it to generate candidate individuals. Unlike traditional LLM-based evolutionary approaches, we leverage the MA-Evolution System to perform both crossover and mutation, enabling more targeted and problem-aware generation of new individuals. Specific prompts are employed only in the first generation during initialization to generate the initial population. In subsequent generations, the initialization step reuses high-quality algorithms obtained from the previous round. Parent heuristics are combined using specific prompts to create new candidate algorithms during crossover. Mutation then slightly alters these candidates to explore nearby solutions. To balance exploration and exploitation during parent selection, we adopt fitness-proportional selection  \citep{zhou2019evolutionary} by assigning selection probabilities to individuals based on their fitness scores.

\textbf{MA-Evolution System} To generate high-quality heuristics, we propose a multi-agent evolution system inspired by multi-agent system \citep{mad1,mad2,mad3,mad4}. As shown in \autoref{ma-e}, the process includes three stages. In the first stage, the \textit{Designer} agent receives the MILP task context, existing code, and the specified evolutionary operation. It produces a high-level design plan and procedural outline for a new heuristic. In the second stage, the \textit{Coder} agent implements the algorithm based on the Designer’s plan. The \textit{Reviewer} agent then checks the code by compiling it and performing logical analysis, providing feedback and suggestions. The \textit{Coder} and \textit{Reviewer} iteratively improve the code through several rounds of interaction. In the final stage, if no consensus is reached, the \textit{Judge} agent reviews the full interaction history and feedback, and makes a final decision on the output code and its description.

\textbf{Prompt engineering}
Our prompts are constructed based on three essential elements: the designated role of the large language model within the MA-E System, contextual information about the MILP problem, and  evolution-specific operations intrinsic to evolutionary computation, such as mutation.
The full prompt information is presented in the appendix.

\section{Experiments}
\subsection{Experimental settings}
 To demonstrate the superiority of our method in the diving task, we conduct three sets of experiments across six MILP datasets. 1) The first set of experiments was designed to study the diving performance of our method and compare it against existing diving heuristics. 
 2) To evaluate the solving efficiency of our method, we conduct a second set of experiments comparing it against default SCIP \citep{achterberg2007constraint}, tuned SCIP, and EoH \citep{liu2024evolution}. 
 3) To evaluate the effectiveness of our method on real-world datasets, we conducted a third set of experiments. We replaced the default diving heuristics in the solver with our proposed method and aimed to improve overall performance on practical MILP tasks. The results of this benchmark experiment are reported in the Appendix.

\subsection{Experiments for diving performance}
\textbf{Experimental setup}
With this first set of experiments, we evaluate a total of 11 publicly available diving methods on four combinatorial optimization benchmarks: cauctions, setcover, facilities, and indset. These heuristics include 6 human-designed algorithms integrated into the open-source solver SCIP (\emph{i.e.}, \textit{coefficient}, \textit{fractional}, \textit{linesearch}, \textit{pseudocost}, \textit{distributional}, \textit{vectorlength}, and \textit{farkas} \citep{witzig2021conflict}), a learning-based GNN method L2DIVE \citep{paulus2023learning}, and 4 heuristics generated via large language models using evolutionary computation. Among the LLM-based methods, we include LLM4Solver \citep{zhou2024llm4solver}, FunSearch \citep{romera2024mathematical}, EoH \citep{liu2024evolution} and HillClimb \citep{HillClimb}.

\begin{table}[ht]
\centering
\small
\caption{The standard error and average relative primal gap of different diving heuristics. The results compare our method with other LLM-based evolutionary approaches, as well as seven human-designed and one learning-based baseline.}
\setlength{\extrarowheight}{0pt}
\addtolength{\extrarowheight}{\aboverulesep}
\addtolength{\extrarowheight}{\belowrulesep}
\setlength{\aboverulesep}{0pt}
\setlength{\belowrulesep}{0pt}

\begin{tabular}{cccccc}
\toprule
\label{table1}
Category & Method & Cauctions & Facilities & Setcover & Indset \\ 
\midrule

\multirow{5}{*}{\makecell[c]{LLM-based\\Evolution}} 
& \cellcolor{gray!15}\textbf{DHEvo(Ours)} & \cellcolor{gray!15}\textbf{1.92 (2.45)} & \cellcolor{gray!15}\textbf{0.70 (1.40)} & \cellcolor{gray!15}\textbf{9.74 (7.35)} & \cellcolor{gray!15}\textbf{1.12 (1.31)} \\

& LLM4Solver \citep{zhou2024llm4solver}         & 2.50 (3.50) & 0.85 (1.42) & 18.33 (19.26) & 1.13 (1.15) \\
& Funsearch \citep{romera2024mathematical}          & 3.04 (7.35) & 1.18 (3.06) & 77.99 (83.89) & 1.61 (3.75) \\
& HillClimb \citep{HillClimb}.          & 6.10 (60.30) & 0.75 (1.40) & 81.55 (343.17) & 1.61 (3.75) \\
& EoH  \citep{liu2024evolution}               & 3.15 (3.15) & 0.80 (1.47) & 20.39 (19.70) & 0.92 (1.06) \\
\midrule

\multirow{6}{*}{\makecell[c]{Hand-crafted\\Heuristics}} 
& Coeficient              & 23.67 (2.14) & 3.20 (3.76) & 68.58 (345.99) & 4.23 (14.42) \\
& Distributional      & 47.80 (71.56) & 1.46 (2.12) & 75.79 (325.90) & 2.57 (10.59) \\
& Farkas            & 23.32 (0.89) & 1.04 (1.64) & 8.13 (8.22)    & - \\
& Pseudocost             & 22.51 (2.30) & 1.06 (1.23) & 23.56 (30.31)  & 3.31 (2.98) \\
& Linesearch        & 22.95 (0.90) & 13.80 (10.94) & 68.59 (346.00) & 3.31 (3.10) \\
& Vectorlength           & 42.93 (83.57) & 13.93 (10.61) & 68.59 (346.01) & 8.89 (7.61) \\
\midrule

Learning-based & L2DIVE \citep{paulus2023learning} & 2.60 & 0.71 & 3.58 & 1.37 \\
\bottomrule
\end{tabular}
\end{table}

To independently investigate the quality of heuristic algorithms, diving is applied solely at the root node of each instance. All other components of the solver, such as branching, cutting planes, and other primal heuristics, are disabled. For fitness evaluation during evolution, we generate 50 instances each for setcover, cauctions, and indset, and 25 instances for facilities. We test the discovered diving heuristics on 100 instances each. For fairness, all LLM-based evolutionary algorithms are trained on the same dataset mentioned above. They also use the same API interfaces. Moreover, we ensure that the prompts used for these methods are aligned in terms of task context, including MILP-specific background and diving-related objectives, to match the design of our proposed method.

\textbf{Experimental results}
We compare our method to various baselines in \autoref{table1}. Under the metric of average primal gap, our method consistently achieves strong results across all datasets. In particular, on the Indset dataset, our approach improves over the best manually designed heuristic by 56.04\%. Compared to other LLM-based algorithm design methods, our approach also achieves state-of-the-art performance. For example, on the Setcover dataset, our method surpasses the best LLM-based baseline by 61.8\%. More importantly, in terms of performance variance, our method achieves the lowest variance across all four datasets. Notably, on the setcover dataset, our approach reduces variance by 46.9\% compared to the best-performing LLM-based algorithm design method.

Considering both average performance and variance, our method consistently outperforms all baselines across the four combinatorial optimization datasets. This demonstrates the strong effectiveness and robustness of our approach in generating high-quality heuristics that generalize well across diverse problem instances.

\subsection{Experiments for solving efficiency}
\textbf{Experimental setup}
To compare the solving efficiency of our method, we use three baselines: default SCIP, tuned SCIP (with adjusted freq and freqofs), and EoH. We run experiments on the same four combination optimization benchmark datasets. For each dataset, we randomly generate 1000 instances and select the 100 most challenging ones for testing.  We set a time limit of \( T_{\text{limit}} = 900 \) seconds for each instance and use the primal-dual integral to evaluate the solving efficiency.

\textbf{Experimental results}
We compare our method to the default, tuned SCIP settings and EoH, as shown in \autoref{sloving-time}. Results demonstrate that our approach not only improves solution quality but also leads to better solving efficiency. On the challenging facility dataset, our method outperforms the current state-of-the-art by 6.7\% in solving time and 2.8\% in primal-dual integral.

\begin{table}
\centering
\small
\caption{Performance comparison of our method, EoH, default SCIP and tuned SCIPP. Each cell reports the solving time
and the primal-dual integral (PDI) as time / PDI.}
\label{sloving-time}
\setlength{\extrarowheight}{0pt}
\addtolength{\extrarowheight}{\aboverulesep}
\addtolength{\extrarowheight}{\belowrulesep}
\setlength{\aboverulesep}{0pt}
\setlength{\belowrulesep}{0pt}
\begin{tabular}{ccccc} 
\toprule
Method& Cauctions           & Facility                                          & Setcover            & Indset                 \\ 
\hline
Default SCIP~                                   & 4.08/55.87          & 301.20/506.71                                     & 2.43/117.65         & 21.07/230.33           \\
Tuned SCIP                                      & 2.73/24.21          & 201.64/553.15                                     & 2.33/77.02          & 22.71/167.43           \\
EoH \citep{liu2024evolution}                                             & 2.62/37.12          & 197.35/504.56                 & 2.76/96.75          & 20.32/151.34           \\
\rowcolor{gray!15} \textbf{DHEvo(Ours)} & \textbf{2.28/23.42} & \textbf{\textbf{\textbf{\textbf{181.27/490.43}}}} & \textbf{2.27/75.88} & \textbf{18.54/146.39}  \\
\bottomrule
\end{tabular}
\vspace{-3mm}
\end{table}

\subsection{Ablation Study}

To verify the effectiveness of each component in our method, we conduct an ablation analysis of our method using four combinatorial optimization datasets. Our framework contains two main components: data-algorithm co-evolution and multi-agent evolution system. Next, we will conduct ablation experiments on these two parts to verify their effectiveness on the validation dataset.

\noindent\textbf{Analysis on data-algorithm co-evolution} We conduct a comprehensive ablation study and comparative evaluation to assess the effectiveness of the data-algorithm co-evolution mechanism. When this mechanism is excluded, the fitness score is computed as the average relative primal gap across all training instances without any selection. As a result, the variance in the performance of evolved heuristics increases by approximately 10\% on each dataset. Meanwhile, the overall solution quality declines by around 10\% across all datasets. This degradation arises because all training instances are treated equally during evolution, allowing complex or unrepresentative instances to dominate the optimization process, which undermines the generalization of the learned heuristics. Additionally, we replace our MA-Evolution System with the existing best LLM-based evolutionary methods EoH. As shown in \autoref{tab:ablation-mae}, the variants augmented with our co-evolution framework achieve markedly better generalization performance compared to their original versions. In particular, the variance of the evolved heuristics decreases by nearly 30\% on the setcover dataset when the co-evolution mechanism is removed.

\begin{table}
\centering
\small
\setlength{\extrarowheight}{0pt}
\addtolength{\extrarowheight}{\aboverulesep}
\addtolength{\extrarowheight}{\belowrulesep}
\setlength{\aboverulesep}{0pt}
\setlength{\belowrulesep}{0pt}
\caption{Comparison of standard error and average relative primal gap on validation dataset, including our full method, its variant without co-evolution, EoH baseline, and EoH with co-evolution.}
\label{tab:ablation-mae}
\begin{tabular}{ccccc} 
\toprule
                    Method                      & Cautions & Facilities & Setcover     & Indset      \\ 
\midrule
 \rowcolor{gray!15}  \textbf{DHEvo}   & \textbf{2.15(2.53)}    & \textbf{0.83(1.30)} & \textbf{9.74(13.42)}  & \textbf{1.01(1.03)}  \\
DHEvo\_OFF                                & 2.33(2.79)    & 0.93(1.45) & 10.8(13.99)  & 1.23(1.11)  \\
\rowcolor{gray!15}  \textbf{EoH\_DH} & \textbf{2.90(5.60)}    & \textbf{0.84(1.47)} & \textbf{18.31(17.48)} & \textbf{1.07(1.14)}  \\
EoH                                       & 4.38(6.15)    & 1.96(4.36) & 26.14(28.89) & 1.36(1.21)  \\
\bottomrule
\end{tabular}
\vspace{-3mm}
\end{table}

\noindent\textbf{Analysis on MA-Evolution System}
To verify the effectiveness of the MA-Evolution System in generating higher-quality diversity generated individual algorithms, we conduct an ablation study by removing this system from our framework and comparing it with the original version in the setcover dataset.

\begin{table}[ht]
\centering
\begin{minipage}[t]{0.48\textwidth}

To evaluate the diversity of algorithms generated by the MA-Evolution System, inspired by diversity indicator metrics~ \citep{diversity1,diversity2}, we introduce a diversity index defined as
\[
DI = \frac{H}{\log_2 N},
\]
where \( H \) is the Shannon entropy of the score distribution over \( N \) generated samples. A value closer to 1 indicates higher diversity among solutions.
\end{minipage}
\hfill
\begin{minipage}[t]{0.48\textwidth}
\centering
\caption{Ablation study of the MA-Evolution system in terms of average primal gap (APG), diversity index (DI), and primal gap standard deviation (PGSD).}
\label{abmase}
\setlength{\extrarowheight}{0pt}
\addtolength{\extrarowheight}{\aboverulesep}
\addtolength{\extrarowheight}{\belowrulesep}
\setlength{\aboverulesep}{0pt}
\setlength{\belowrulesep}{0pt}

\begin{tabular}{cccc}
\toprule
Method & APG & DI & PGSD \\
\midrule
MA-Evolution OFF & 9.14 & 0.76 & 8.75 \\
 \rowcolor{gray!15}  
\textbf{MA-Evolution ON} & \textbf{8.00} & \textbf{0.88} & \textbf{4.78} \\
\bottomrule
\end{tabular}
\end{minipage}
\end{table}

As shown in the \autoref{abmase}, the algorithms generated by the MA-Evolution System achieve significantly lower average primal gaps, improving by 12.4\% compared to those without the MA-Evolution System. Additionally, they show a 15.8\% improvement in the diversity index, demonstrating the superior diversity of the generated heuristics.

\section{Conclusion}

We present a novel data-algorithm co-evolution framework for solving MILP. By iteratively identifying the most representative instances and co-evolving heuristic algorithms based on them, our method significantly improves the generalization ability of the generated heuristics within the same problem class. Unlike traditional approaches that treat training data as static, our method selects representative  instances during the evolutionary process, enabling the algorithm to generalize better across diverse problem distributions. Meanwhile, heuristic algorithms are co-evolved with the instance set, ensuring mutual adaptation and continuous improvement. We also introduce a multi-agent evolutionary system to improve generation quality and solution diversity. Experimental results show that our approach significantly outperforms existing human-designed, learning-based, and LLM-based baselines in both the primal gap and solving efficiency.

\bibliographystyle{unsrt}
\bibliography{main}

\newpage
\appendix
\section{Diving Heuristics}
\label{appendix:diving_heuristics}

Diving heuristics are primal heuristics that iteratively fix variables based on LP relaxation solutions, simulating a depth-first search in the branch-and-bound tree. Given the LP relaxation of an MILP:
\[
z_{LP}^{\dagger} := \min_{x \in P_{LP}^{\dagger}} c^{\top}x, \quad P_{LP}^{\dagger} = \left\{ x \in \mathbb{R}^{n} \mid Ax < b, \underline{\pi} \leq x \leq \overline{\pi} \right\},
\]
the algorithm starts from an LP solution \( \hat{x} \in P_{LP}^{\dagger} \) and incrementally fixes fractional variables \( x_j \notin \mathbb{Z} \) to integer values. At each step, the feasible region is updated with new bound constraints, and the relaxed problem is re-solved. This process emulates a depth-first traversal of the search space, aiming to quickly construct a feasible integer solution. In general, a generic diving heuristic can be described by Algorithm~\ref{alg:diving}. The only difference among various diving heuristics lies in the scoring function \( s(\cdot) \), which determines the variable to round and the direction of rounding at each iteration.

\begin{algorithm}
\caption{Generic Diving Heuristic}
\label{alg:diving}
\begin{algorithmic}[1]
\Input{MILP with relaxed feasible region $P^*$, LP solution $x^*$, maximum depth $d_{\text{max}}$}
\Output{A set of feasible solutions $\mathcal{X}$ (if found)}
\Require{A scoring function $s$ for selecting branching variables and their rounding direction}
\State Initialize depth $d \gets 1$, candidate set $\mathcal{C} \gets \{j \in \mathcal{I} \mid x^*_j \notin \mathbb{Z}\}$
\While{$d \leq d_{\text{max}}$}
    \State $j \gets \arg\max_{i \in \mathcal{C}} s(x_i)$
    \If{round up}
        \State $l_j \gets \lceil x^*_j \rceil$
    \Else
        \State $u_j \gets \lfloor x^*_j \rfloor$
    \EndIf
    \State $P^* \gets P^* \cap \{l_j \leq x_j \leq u_j\}$
    \If{$P^*$ is infeasible}
        \State \textbf{break}
    \EndIf
    \State $x^* \gets \arg\min_{x \in P^*} c^\top x$
    \If{$x^*$ is roundable}
        \State $\mathcal{X} \gets \mathcal{X} \cup \text{round}(x^*)$
    \EndIf
    \State $d \gets d + 1$
    \State Update candidate variable index set $\mathcal{C}$
\EndWhile
\end{algorithmic}
\end{algorithm}

Here are some diving heuristic algorithms included in SCIP.

\textbf{Coefficient.}  
This strategy selects a variable that has the smallest number of positive up-locks or down-locks. These locks represent how many constraints would prevent increasing or decreasing the variable, respectively. The variable is then fixed in the direction where fewer locks occur. If there is a tie between multiple variables, the method uses a secondary rule called fractional diving to break the tie.

\textbf{Distribution.}  
This method is based on the empirical distribution of fractional values observed in historical solutions. It favors variables that are more frequently fractional in previous LP relaxations. The idea is that such variables are likely to remain fractional and therefore more useful for branching. 

\textbf{Farkas.}  
This strategy tries to construct a Farkas proof to show the infeasibility of the current LP relaxation after branching. It selects the variable whose rounding, in the direction that improves the objective, is predicted to cause the largest gain. This prediction is based on LP dual information or inference from constraint violation. The method is designed to make branching decisions that quickly lead to pruning.

\textbf{Fractional.}  
This method selects the variable that is closest to an integer value, but still fractional. The measure used is \( \left| x_j^* - \lfloor x_j^* + 0.5 \rfloor \right| \), which captures how far the variable's value is from the nearest integer. The selected variable is then rounded in the direction that brings it closest to an integer. This approach is simple and focuses on reducing the integrality gap.

\textbf{Linesearch.}  
This method traces a straight line (ray) from the root node LP solution to the current LP solution \( x^* \). It identifies which integer hyperplane—either \( x_j = \lfloor x_j^* \rfloor \) or \( x_j = \lceil x_j^* \rceil \)—is intersected first along this ray. The variable defining that hyperplane is selected for branching. This approach can be seen as a geometric way to decide which variable will influence the search path as soon as possible.

\textbf{Pseudocost.}  
This strategy uses historical data, called pseudocosts, to guide branching. For each variable, it records the average objective improvement caused by previous up- or down-branching decisions. The variable and branching direction with the highest expected improvement are selected. This method also considers the current fractionality of the variable to refine the choice. It is widely used due to its balance between accuracy and efficiency.

\textbf{Vectorlength.}  
This method is inspired by set-partitioning problems. It evaluates the trade-off between how much rounding a variable is expected to degrade the objective, and how many constraints the variable appears in. The selected variable minimizes the ratio between the expected degradation and its constraint count. This helps prioritize variables that have a broad structural impact while limiting damage to the objective.

To guide our learned diving score function, we use variable-level features that are inspired by those employed in existing human-designed diving heuristics. These include 13 features in total, which are listed and described in \autoref{13feature}.

\begin{table}[ht]
\centering
\small
\caption{Description of the 13 input features used in the diving score function.}
\label{13feature}
\begin{tabular}{lp{11cm}}
\toprule
\textbf{Feature Name} & \textbf{Feature Description} \\
\midrule
\textit{mayrounddown} & Boolean; indicates whether the variable can be rounded down while maintaining feasibility. \\
\textit{mayroundup} & Boolean; indicates whether the variable can be rounded up while maintaining feasibility. \\
\textit{candsfrac} & Float; fractional part of the variable’s value in the LP relaxation, i.e., \( |x_j^* - \lfloor x_j^* \rfloor| \). \\
\textit{candsol} & Float; value of the variable in the current LP relaxation solution. \\
\textit{nlocksdown} & Integer; number of down-locks, i.e., constraints that would be violated by decreasing the variable. \\
\textit{nlocksup} & Integer; number of up-locks, i.e., constraints that would be violated by increasing the variable. \\
\textit{obj} & Float; coefficient of the variable in the objective function. \\
\textit{objnorm} & Float; Euclidean norm of the objective function coefficient vector. \\
\textit{pscostdown} & Float; pseudocost for decreasing the variable's value. \\
\textit{pscostup} & Float; pseudocost for increasing the variable's value. \\
\textit{rootsolval} & Float; value of the variable in the LP relaxation at the root node. \\
\textit{nNonz} & Integer; number of nonzero entries in the variable's column in the constraint matrix. \\
\textit{isBinary} & Boolean; TRUE if the variable is binary, i.e., has domain \(\{0,1\}\). \\
\bottomrule
\end{tabular}
\label{tab:features}
\end{table}

\section{More experiment details}
In all the experiments, we evaluate the performance of agents driven by GPT-4o mini across various tasks. We run all the experiments with three random seeds on Intel(R) Xeon(R) CPU E5-2667 v4 @ 3.20GHz and NVIDIA A100. 

\textit{Note:} Since the code for L2DIVE is currently not open-source and specific hyperparameters are unavailable, we
officially report the performance of L2DIVE based on its ratio to the best human-designed heuristic as presented in the original article.
\subsection{SCIP settings}
To construct our Tuned baseline, we incorporated domain knowledge and performed a randomized search over key diving-related parameters in SCIP 7.0.2. The primary parameters that govern the invocation of individual diving heuristics are \textit{freq} and \textit{freqofs}. These parameters determine when and how frequently a given diving heuristic is triggered during the branch-and-bound process. By adjusting their values, we can generate diverse solver behaviors that vary the timing and intensity of heuristic application. For each diving heuristic, we independently sampled its configuration by setting \textit{freq} to one of four values with equal probability: \(-1\) (disabled), \(\lfloor 0.5 \times \textit{freq}_{\text{default}} \rfloor\) (increased frequency), \(\textit{freq}_{\text{default}}\) (default frequency), or \(\lfloor 2 \times \textit{freq}_{\text{default}} \rfloor\) (reduced frequency). In parallel, we randomly set \textit{freqofs} to either zero or its default value, also with equal probability. This approach allows us to sample a wide range of heuristic schedules while maintaining compatibility with established SCIP parameter semantics.

\subsection{Datasets}

We evaluate our method on seven benchmark datasets, including four synthetic combinatorial optimization problems and three real-world MILP tasks. The datasets are widely used in prior work and include:

\begin{itemize}
   \item \emph{Setcover}: A classical combinatorial problem where the objective is to select a minimum number of subsets such that their union covers all elements. Instances are represented as binary matrices with rows corresponding to elements and columns to subsets. Easy instances have 500 rows and 1000 columns, while hard instances increase the size to 2000 rows and 1000 columns.
    \item \emph{Cauctions}: A combinatorial auction problem where bidders submit bids on bundles of items, aiming to maximize total revenue without violating item availability constraints. Easy instances contain 100 items and 500 bids, while hard instances include 300 items and 1500 bids.
    \item \emph{Facilities}: A capacitated facility location problem involving the selection of facility sites and the assignment of customers to minimize facility opening and service costs. Easy instances consist of 100 facilities and 100 customers, whereas hard instances have 100 facilities and 400 customers.
    \item \emph{Indset}: The maximum independent set problem, which seeks the largest possible set of mutually non-adjacent vertices within a graph. Easy instances feature 500 nodes with an affinity of 4, and hard instances have 1500 nodes with the same affinity.
    \item \emph{LoadBalance}: A server load balancing problem arising in distributed systems, modeled as an MILP.
    \item \emph{NNVerify}: A verification problem for neural networks, where constraints encode input-output relationships that must be satisfied.
    \item \emph{MIPLIB}: It contains a diverse collection of real-world and academic instances spanning various domains such as scheduling, network design, logistics, and combinatorial optimization. We selected 20 instances for experimental comparison.

\end{itemize}

The first and second experimental groups are conducted on the four synthetic datasets, focusing on diving performance and solving efficiency. The third group uses the three real-world datasets to demonstrate the effectiveness of our method in practical applications.

\begin{table}
\centering
\small
\caption{Table 10: Used MIPLIB instance names}
\label{tab:miplib_instances}
\begin{tabular}{lllll} 
\toprule
air05        & beasleyC3       & binkar10\_1        & cod105          & dano3\_3     \\
eil33-2      & hypothyroid-k1  & istanbul-no-cutoff & markshare\_4\_0 & mas76        \\
mc11         & mik-250-20-75-4 & n5-3               & neos-860300     & neos-957323  \\
neos-1445765 & nw04            & piperout-27        & pk1             & seymour1     \\
\bottomrule
\end{tabular}
\end{table}

\section{Experiments for practical MILP tasks.}
\subsection{Experimental settings}
To evaluate the effectiveness and generalization ability of the proposed heuristic framework, we conduct experiments on three representative datasets: LoadBalance, MILPLIB, and NNVerify, which cover a broad range of MILP problem structures. Across all datasets, we adopt two standard performance metrics: the \emph{primal-dual integral} (PDI), which captures convergence behavior and solution quality over time, and the \emph{solving time} (T), which measures how quickly a feasible or optimal solution is found. For LoadBalance, we use 100 instances for validation and another 100 for testing, with \( T_{\text{limit}} = 3600 \) seconds as the standard setting and \( T_{\text{limit}} = 1800 \) seconds for additional robustness evaluation under tighter budgets. For MILPLIB, we select 20 relatively simple benchmark instances as a test set to evaluate generalization performance on classical MILP formulations; the instance names are listed in \autoref{tab:miplib_instances}. For NNVerify, we evaluate on 100 testing instances derived from neural network verification problems, using a time limit of \( T_{\text{limit}} = 900 \) seconds and considering only instances successfully solved within the limit. To isolate the contribution of the learned diving heuristics, we perform all experiments under both \texttt{cut-selection} enabled and disabled configurations. In all settings, the heuristics are integrated into SCIP, and the best-performing variant is selected on the validation set based on either PDI or solving time before being applied to the testing set, mirroring realistic deployment scenarios.

\subsection{Experimental results}
We compare our method to the default  SCIP settings, as shown in \autoref{tab:comparison-hard}. The experimental results demonstrate that our method achieves competitive improvements across all datasets. With cut plane selection enabled, our method achieves a 26.1\% improvement in the primal dual integral (PDI) on the LoadBalance dataset. On the NNVerify dataset, our approach more than doubles the solving efficiency regardless of whether cut selection is enabled. For the MIPLIB dataset, our method improves solving efficiency by 36\% and achieves a 12\% reduction in PDI compared to the baseline.

\begin{table}
\centering
\small
\setlength{\extrarowheight}{0pt}
\addtolength{\extrarowheight}{\aboverulesep}
\addtolength{\extrarowheight}{\belowrulesep}

\caption{Comparison of solving time and primal-dual integral (PDI) across different methods and datasets. The suffix \textit{-off-cut} indicates that cut plane selection is disabled, while \textit{-on-cut} means cut plane selection is enabled. All methods fail to obtain the optimal solution on the Load balancing dataset within the time limit(3600 or 1800).}
\label{tab:comparison-hard}
\begin{tabular}{lcccccc} 
\toprule
& \multicolumn{2}{c}{Load balancing} & \multicolumn{2}{c}{Neural network verif} & \multicolumn{2}{c}{MIPLIB}  \\ 
\cmidrule(lr){2-3} \cmidrule(lr){4-5} \cmidrule(lr){6-7}
& Solving time  & PDI        & Solving time  & PDI                      & Solving time & PDI          \\ 
\midrule
\rowcolor{gray!15} Ours-off-cut & 3600  & 346980.53~ & 72.42         & 5413.32                  & 263.48        & 12101.11     \\
Scip-off-cut                    & 3600  & 347597.70  & 669.15        & 38455.17                 & 469.22        & 18127.57     \\
\rowcolor{gray!15} Ours-on-cut & 1800  & 7305.2     & 35.67         & 2744.21                  & 117.67        & 5599.62      \\
Scip-on-cut                    & 1800  & 9881.29    & 137.19        & 8210.46                  & 184.3         & 6339.64      \\
\bottomrule
\end{tabular}
\end{table}

\section{Prompts}
\subsection{Prompts design}
\begin{figure}
  \centering
  \small
\includegraphics[width=\textwidth]{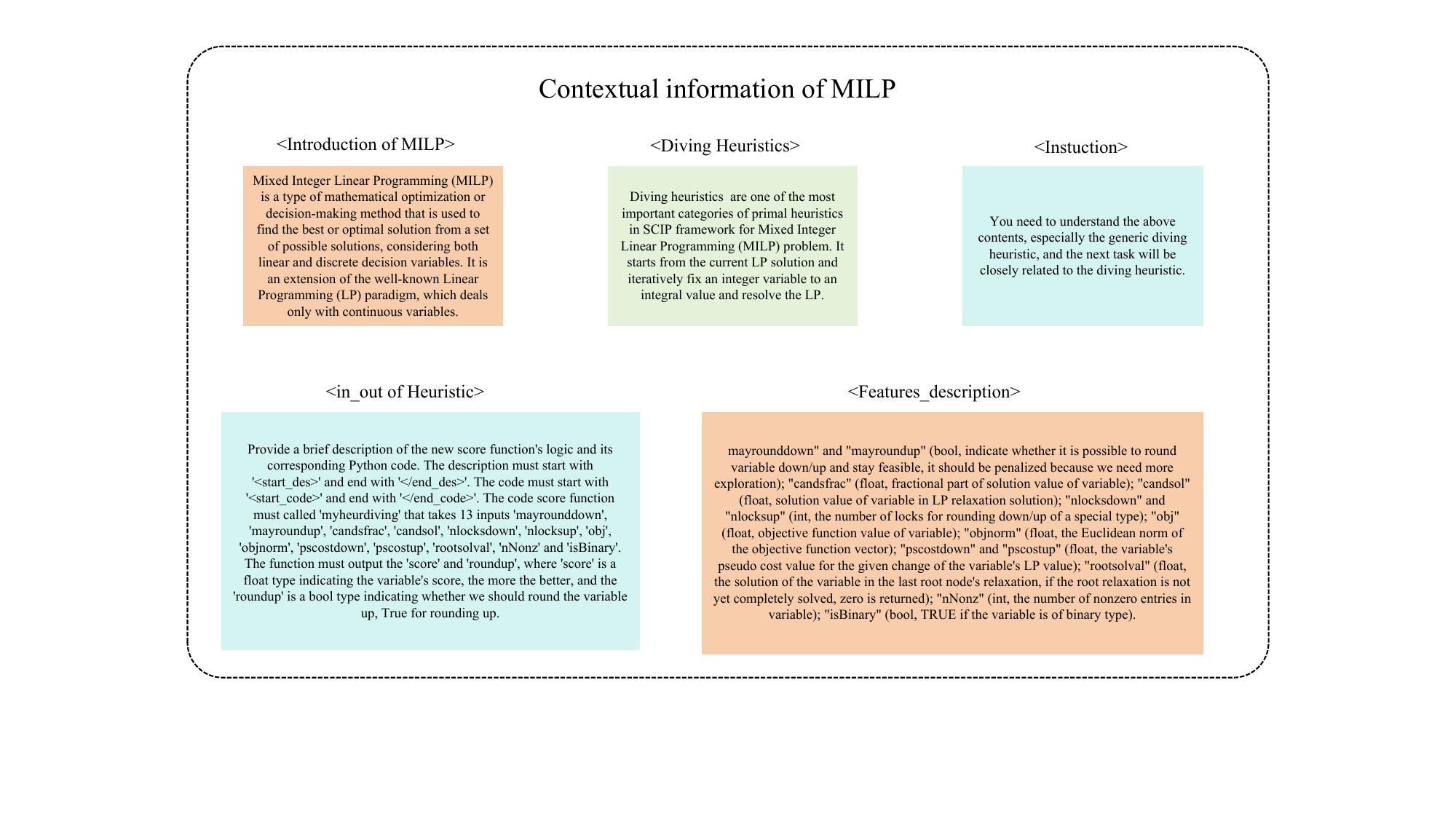}
    \caption{The prompts of contextual information of MILP.}
  \label{milp-app}

\end{figure}
Our prompt design adopts a structured and modular format to effectively guide LLMs in performing evolutionary search within the multi-agent evolutionary framework. Each prompt is composed of three essential components, designed to provide the LLM with both domain-specific grounding and a clear operational goal.

\begin{figure}
  \centering
  \small
\includegraphics[width=\textwidth]{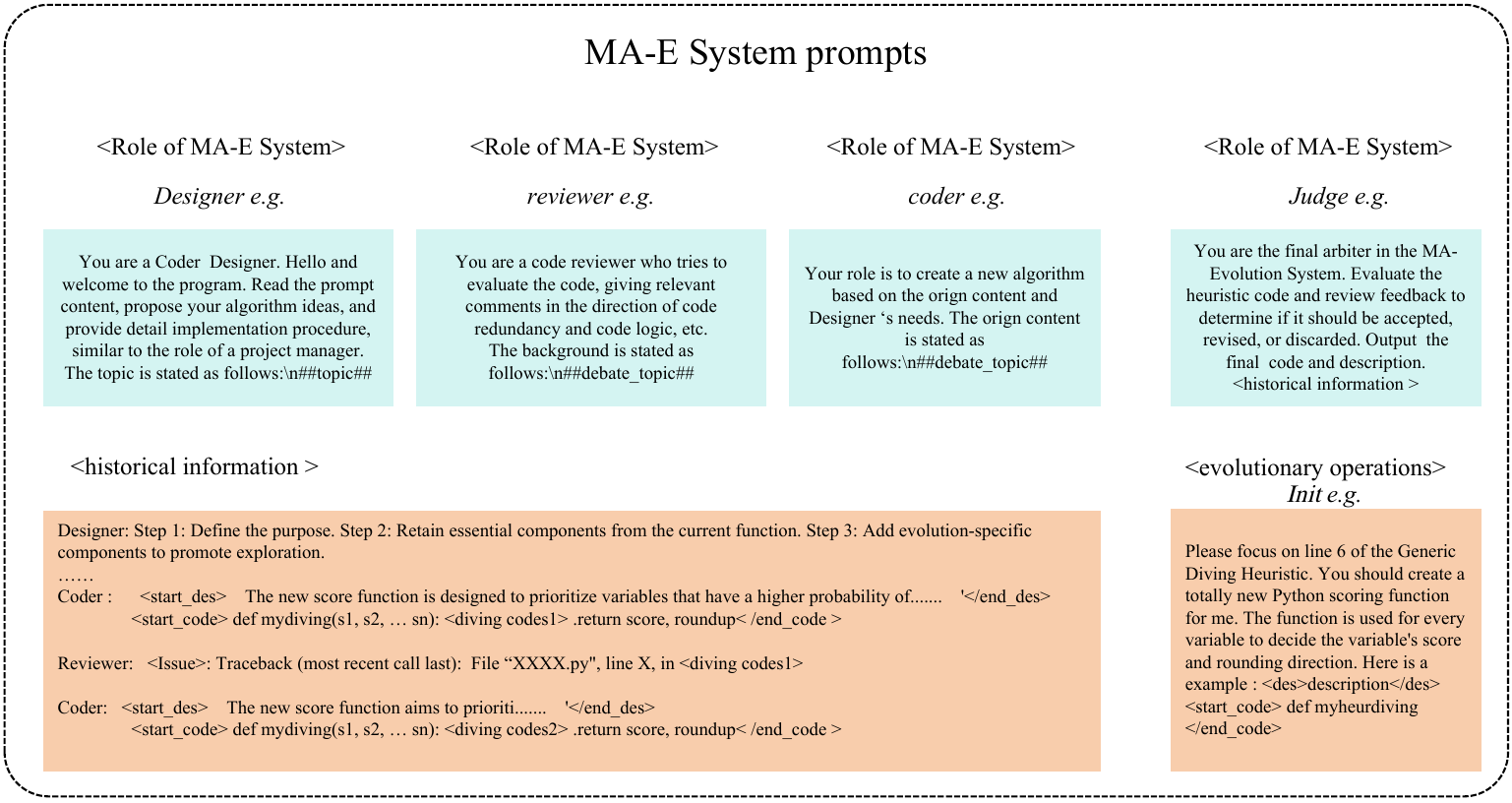}
    \caption{The prompts in MA-Evolution System.}
  \label{mae-app}

\end{figure}

As shown in \autoref{milp-app}, background prompts contain \textit{<Introduction of MILP>, <Diving Heuristics>, <Instuction>
, <in\_out of Heuristic>, <Features description>}. Together,
they provide enough background knowledge of diving heuristics for the downstream tasks. 
Prompts in MA-Evolution System are modular and follow a structured template to ensure consistency across generations, as shown in \autoref{mae-app}. At the core of each prompt are three elements: (1) the functional role of the agent, which defines the nature of the task (\emph{e.g.}, proposing a new heuristic or reviewing existing code); (2) a formal or semi-formal description of the MILP problem to ground the response in the relevant optimization context; and (3) a specification of the evolutionary operation that informs the agent's goal in the current generation cycle.

Our evolution operation's prompt includes three main types: initialization, mutation, and crossover. Each type corresponds to a distinct stage in the evolutionary search process and is designed to guide LLMs in generating or improving heuristics for MILP diving.

\textbf{Initialization}  
The LLM is instructed to create a new scoring function from scratch. The function should assign a score and a rounding direction to each fractional variable, based only on the LP relaxation and objective function. This stage initializes the population with diverse and problem-aware heuristics.

\textit{Example prompt:} Please create a new Python scoring function for a Generic Diving Heuristic. The function should assign a score and rounding direction to each fractional variable, using only information from the LP relaxation and the objective function.

\textbf{Mutation}  
The LLM receives an existing scoring function and modifies it slightly. The modification should be meaningful and aimed at improving performance or exploring nearby variants in the heuristic space. This enables local search around known good solutions.

\textit{Example prompt:} Please make a small but meaningful change that may improve performance or explore alternative behavior. Ensure the result is syntactically correct and remains within the MILP context.\\
Original function: [insert code]

\textbf{Crossover}  
The LLM combines two existing scoring functions into a new one. It should preserve useful components from both parents while ensuring the resulting function is coherent and consistent. This enables global search by recombining successful patterns.

\textit{Example prompt:} You are creating a new heuristic by combining two existing ones. Please synthesize a scoring function that inherits effective components from both parents while maintaining logical consistency.\\
Heuristic A: [insert code]\\
Heuristic B: [insert code]

\subsection{Example}
The following is an example of our method applied within DHEvo:

\textbf{Designer} You are a Coder Designer. Hello and welcome to the program. Read the prompt content, propose your algorithm ideas, and provide detail implementation procedure, similar to the role of a project manager.
The topic is stated as follows:Diving heuristics  are one of the most important categories of primal heuristics in SCIP framework for Mixed Integer Linear Programming (MILP) problem. It starts from the current LP solution and iteratively fix an integer variable to an integral value and resolve the LP. You should create a totally new Python scoring function for me (different from the heuristics in the literature) to choose the fractional variable and corresponding rounding direction using the information of the LP relaxation and objective function. The function is used for every variable to decide the variable's score and rounding direction. Specifically, you have these features to use in the score function: "mayrounddown" and "mayroundup" (bool, indicate whether it is possible to round variable down/up and stay feasible, it should be penalized because we need more exploration); "candsfrac" (float, fractional part of solution value of variable); "candsol" (float, solution value of variable in LP relaxation solution); "nlocksdown" and "nlocksup" (int, the number of locks for rounding down/up of a special type); "obj" (float, objective function value of variable); "objnorm" (float, the Euclidean norm of the objective function vector); "pscostdown" and "pscostup" (float, the variable's pseudo cost value for the given change of the variable's LP value); "rootsolval" (float, the solution of the variable in the last root node's relaxation, if the root relaxation is not yet completely solved, zero is returned); "nNonz" (int, the number of nonzero entries in variable); "isBinary" (bool, TRUE if the variable is of binary type).
Provide a brief description of the new score function's logic and its corresponding Python code. The description must start with '<start\_des>' and end with '</end\_des>'. The code must start with '<start\_code>' and end with '</end\_code>'. The code score function must call 'myheurdiving' that takes 13 inputs 'mayrounddown', 'mayroundup', 'candsfrac', 'candsol', 'nlocksdown', 'nlocksup', 'obj', 'objnorm', 'pscostdown', 'pscostup', 'rootsolval', 'nNonz', and 'isBinary'. The function must output the 'score' and 'roundup', where 'score' is a float type indicating the variable's score, the more the better, and the 'roundup' is a bool type indicating whether we should round the variable up, True for rounding up. Be creative and do not give additional explanations.

\textbf{Coder} Your role is to create a new algorithm based on the original content and Designer‘s needs. The original content is stated as follows: Diving heuristics are one of the most important categories of primal heuristics in the SCIP framework for Mixed Integer Linear Programming (MILP) problems. It starts from the current LP solution and iteratively fixes an integer variable to an integral value and resolves the LP.
 You should create a new Python scoring function for me (different from the heuristics in the literature) to choose the fractional variable and corresponding rounding direction using the information of the LP relaxation and objective function. The function is used for every variable to decide the variable's score and rounding direction.
Specifically, you have these features to use in the score function: 
"mayrounddown" and "mayroundup" (bool, indicate whether it is possible to round variable down/up and stay feasible, it should be penalized because we need more exploration); "candsfrac" (float, fractional part of solution value of variable); "candsol" (float, solution value of variable in LP relaxation solution); "nlocksdown" and "nlocksup" (int, the number of locks for rounding down/up of a special type); "obj" (float, objective function value of variable); "objnorm" (float, the Euclidean norm of the objective function vector); "pscostdown" and "pscostup" (float, the variable's pseudo cost value for the given change of the variable's LP value); "rootsolval" (float, the solution of the variable in the last root node's relaxation, if the root relaxation is not yet completely solved, zero is returned); "nNonz" (int, the number of nonzero entries in variable); "isBinary" (bool, TRUE if the variable is of binary type).
Provide a brief description of the new score function's logic and its corresponding Python code. The description must start with '<start\_des>' and end with '</end\_des>'. The code must start with '<start\_code>' and end with '</end\_code>'. The code score function must call 'myheurdiving' that takes 13 inputs: 'mayrounddown', 'mayroundup', 'candsfrac', 'candsol', 'nlocksdown', 'nlocksup', 'obj', 'objnorm', 'pscostdown', 'pscostup', 'rootsolval', 'nNonz', and 'isBinary'. The function must output the 'score' and 'roundup', where 'score' is a float type indicating the variable's score, the more the better, and the 'roundup' is a bool type indicating whether we should round the variable up, True for rounding up. Be creative and do not give additional explanations. Designer idea: Allow weighting coefficients for each component.

\textbf{Reviewer} You are a code evaluator who tries to evaluate the code, giving relevant comments in the direction of code redundancy and code logic, etc. <code>

\textbf{Judger} 
You are the final arbiter in the MA-Evolution System. Evaluate the
heuristic code and review feedback to determine if it should be accepted,
revised, or discarded. Output the final code and description.
<historical information >

\section{Generated heuristics}
This section presents the best heuristics generated by DHEvo.

\lstset{
  basicstyle=\ttfamily\footnotesize,
  breaklines=true,
  frame=single,
  backgroundcolor=\color{gray!5},
  captionpos=b,
  language=Python,
  keywordstyle=\color{blue},
  commentstyle=\color{gray!70},
  stringstyle=\color{red},
  showstringspaces=false,
  columns=fullflexible
}

\begin{lstlisting}[language=Python, caption={Heuristic for cauctions}]
def myheurdiving(mayrounddown, mayroundup, candsfrac, candsol,
                 nlocksdown, nlocksup, obj, objnorm,
                 pscostdown, pscostup, rootsolval,
                 nNonz, isBinary):
    score = 0.0
    roundup = False

    # Penalize if both rounding options are feasible
    if mayrounddown and mayroundup:
        score = -40

    # Evaluate candidate based on fractional part
    if candsfrac > 0.5:
        score += candsfrac * 80
        roundup = True
        if pscostup > 0.5:
            score += pscostup * 50
    else:
        score += (1 - candsfrac) * 60
        if pscostdown < -0.3:
            score -= abs(pscostdown) * 25

    # Normalize objective contribution
    score += (obj / (objnorm + 1e-6)) * 90

    # Adjust for locking counts
    score += (nlocksdown * 25 - nlocksup * 15)

    # Reward for non-zero entries and binary variable nature
    if nNonz > 2:
        score += nNonz * 20
    if isBinary:
        score += 50

    return score, roundup
\end{lstlisting}

\begin{lstlisting}[language=Python, caption={Heuristic for facility}]
def myheurdiving(mayrounddown, mayroundup, candsfrac, candsol, nlocksdown, nlocksup, obj, objnorm, pscostdown, pscostup, rootsolval, nNonz, isBinary):
    score = 0.0
    roundup =  False

    # Base score weighted by normalized objective contribution
    score += (obj/(objnorm + 1e-9)) * 5 if objnorm >0 else 0

    # Penalize rounding options to encourage exploration
    score -= nlocksdown * 7 if mayrounddown else 0
    score -= nlocksup *7 if mayroundup else 0

    # Favor large fractions away from 0.5 for exploration
    score += (abs(candsfrac - 0.5) * 10)

    # Adjust score based on solution value and its contribution
    score += (candsol / (1 + abs(rootsolval) * obj)) * 4 if rootsolval != 0 else 0

    # Employ pseudo costs to influence rounding decisions
    if pscostdown < 0 and mayrounddown:
        score += -pscostdown * 3 # Favor rounding down with negative pseudo costs
    if pscostup < 0 and mayroundup:
        score -= -pscostup * 3 # Discourage rounding up with negative pseudo costs

    #Determine rounding direction based on fractional part and exploration potential
    if candsfrac >= 0.7 and mayroundup:
        roundup = True
    elif candsfrac <= 0.3 and mayrounddown:
        roundup = False

    # Encourage solutions with fewer nonzero entries 
    score += (1 / (nNonz + 1)) * 2 if nNonz > 0 else 0

    return score, roundup 

\end{lstlisting}

\begin{lstlisting}[language=Python, caption={Heuristic for indset}]
def myheurdiving(mayrounddown, mayroundup, candsfrac, candsol, nlocksdown, nlocksup, obj, objnorm, pscostdown, pscostup, rootsolval, nNonz, isBinary):
    score = 0.0

    # Strongly penalize feasible rounding options 
    if mayrounddown:
        score -= 3.0
    if mayroundup:
        score -= -3.0
    
    # Incorporate fractional part and objective value
    score += (1.0 - candsfrac) * obj * 0.5 if mayrounddown else 0
    score += candsfrac * obj *0.5 if mayroundup else 0

    # Adjust with pseudo costs
    score += pscostdown * candsfrac * 1.5 if mayrounddown else 0
    score += pscostup * (1 - candsfrac) * 1.5 if mayroundup else 0

    # Apply less severe penalty for distance from the root solution
    score -= abs(rootsolval - candsol) * 0.1

    # Normalize the score
    if objnorm > 0 :
        score /= objnorm
    
    # Reward more for binary variables
    score += nlocksup * 0.3 - nlocksdown * 0.3
    if isBinary:
        score += 1.0

    # Determine rounding direction 
    roundup = (score > 0) and (not isBinary or mayroundup)

    return score, roundup

\end{lstlisting}

\begin{lstlisting}[language=Python, caption={Heuristic for indset}]
def myheurdiving(mayrounddown, mayroundup, candsfrac, candsol, nlocksdown, nlocksup, obj, objnorm, pscostdown, pscostup, rootsolval, nNonz, isBinary):
    score = 0.0

    # Strongly penalize feasible rounding options 
    if mayrounddown:
        score -= 3.0
    if mayroundup:
        score -= -3.0
    
    # Incorporate fractional part and objective value
    score += (1.0 - candsfrac) * obj * 0.5 if mayrounddown else 0
    score += candsfrac * obj *0.5 if mayroundup else 0

    # Adjust with pseudo costs
    score += pscostdown * candsfrac * 1.5 if mayrounddown else 0
    score += pscostup * (1 - candsfrac) * 1.5 if mayroundup else 0

    # Apply less severe penalty for distance from the root solution
    score -= abs(rootsolval - candsol) * 0.1

    # Normalize the score
    if objnorm > 0 :
        score /= objnorm
    
    # Reward more for binary variables
    score += nlocksup * 0.3 - nlocksdown * 0.3
    if isBinary:
        score += 1.0

    # Determine rounding direction 
    roundup = (score > 0) and (not isBinary or mayroundup)

    return score, roundup

\end{lstlisting}

\begin{lstlisting}[language=Python, caption={Heuristic for setcover}]
def myheurdiving(mayrounddown, mayroundup, candsfrac, candsol, nlocksdown, nlocksup, obj, objnorm, pscostdown, pscostup, rootsolval, nNonz, isBinary):
    score = 0.0

    # Penalties for feasible rounding options to promote exploration
    if mayrounddown:
        score -= 10.0
    if mayroundup:
        score -= 10.0
    
    # Favor fractional values at extremes (0 or 1)
    score += (1 - abs(candsfrac - 0.5)) * 30.0

    # Normalize impact of the objective function
    score += (obj/(objnorm + 1e-5)) * 0.5

    #Include pseudo cost adjustments for better decision-making
    score += pscostup if mayroundup else 0.0
    score -= pscostdown if mayrounddown else 0.0

    # Integrate root solution value adjusted by variable complexity
    score += rootsolval / (nNonz + 1)

    # Amplify score for binary variables to encourage decisive rounding
    if isBinary:
        score *= 2.0
    
    # Determine rounding direction based on computed score and pseudo costs
    roundup = (mayrounddown and (pscostup <= pscostdown or not mayrounddown))

    return score, roundup
\end{lstlisting}

\newpage

\end{document}